# Influences Combination Of Multi-Sensor Images On Classification Accuracy♠

Firouz Abdullah Al-Wassai*
Research Student,
Dept. of Computer Science
(SRTMU), Nanded, India
fairozwaseai@yahoo.com

N.V. Kalyankar
Principal,
Yeshwant Mahavidyala College
Nanded, India
drkalyankarnv@yahoo.com

***Abstract:*** This paper focuses on two main issues; first one is the impact of combination of multi-sensor images on the supervised learning classification accuracy using segment Fusion (SF). The second issue attempts to undertake the study of supervised machine learning classification technique of remote sensing images by using four classifiers like Parallelepiped (Pp), Mahalanobis Distance (MD), Maximum-Likelihood (ML) and Euclidean Distance(ED) classifiers, and their accuracies have been evaluated on their respected classification to choose the best technique for classification of remote sensing images. QuickBird multispectral data (MS) and panchromatic data (PAN) have been used in this study to demonstrate the enhancement and accuracy assessment of fused image over the original images using ALwassaiProcess software. According to experimental result of this study, is that the test results indicate the supervised classification results of fusion image, which generated better than the MS did. As well as the result with Euclidean classifier is robust and provides better results than the other classifiers do, despite of the popular belief that the maximum-likelihood classifier is the most accurate classifier.

## 1. INTRODUCTION

The classification is defined as information of extracting process that analyses the adopted spectral signatures by using a classifier and then assigns the spectral vector of pixels to categories according to their spectral. Depending on the level of pattern classification procedure techniques are used in classifying images can be broadly categorized as either supervised or unsupervised. In the case of unsupervised classification means by which pixels in the image are assigned to spectral classes without the user having foreknowledge of training samples or a-prior knowledge of the area. While In the case of supervised classification, it requires the user provide the types of cover sets in the image e.g., water, cobble, deciduous forest, etc. As well as a training field for each cover type. The training field typically corresponds to an area in the image, which has contains of the cover type, and the collection of all training fields is known as the training set or ground-truth data. These training set can be obtained using site visits, maps, aerial photographs or even a photo interpretation of a colour composite product formed from the satellite image data [1]. The ground-truth data is then used to assign each pixel to its most probable cover type. Many factors in every case, the crucial steps are: (i) selection of a set of features which describes the best pattern from the original feature set and thus can be viewed as a principal pre-processing tool prior to solving classification problems [2], (ii) choice of a suitable classifier for the comparison of the pattern describing the object to be classified and the target patterns and (iii) a third stage, that of assessing the degree of accuracy of the allocation process. Many factors affect the accuracy of image classification and the quality of land cover maps, which is often perceived as being insufficient for operational use [3-4]. Classification accuracy is a function of training set selection, and a good training set has the following characteristics [5]:1). It should contain samples describing all classes,2) It should have a sufficient number of independent samples for each class, and 3) It should be made up of samples that completely describe the intra-class variability.

The feature-selection techniques that are most widely used in remote sensing generally require the definition of a discriminant function and a decision rule. The decision rule is a measure of the effectiveness of the considered subset of features, and the discriminant function is an algorithm that aims at efficiently finding a solution (i.e., a subset of features) that optimizes the adopted decision rule. In standard feature-selection methods, several feature-selection algorithms have been proposed for selecting of training set, e.g., [6, 1]. the discriminant functions typically adopted are statistical measures that assess the separability of the different classes on a given training set but do not explicitly take into account the stationarity of the features (e.g., the variability of the spectral signature of the land-cover classes) [7]. This approach may result in selecting a subset of features that retains very good discrimination properties in the portion of the scene close to the training pixels (and therefore with similar behavior), but are not appropriate to model the class distributions in separate portions on the scene, which may present different spectral behavior[8]. In general, Current image processing techniques are limited in their ability to automatically extract accurate land cover features [9].

In the study the developed system User Graphic Interface UGI ALwassaiProcess software was designed to automatic classification by select any number and size of regions that will be the training data of the test image. This is the crucial program for the image of classification, this deals with how to select the training data automatically which describes the best pattern and by this way allow us to determine the interesting class of user of image. The program offers the selections of any size of the training data; it means that the user can decide the increase of the successful of classification by this experiment. Also, in this paper there are comparisons of various classifiers which have been discussed with their accuracies evaluated on their respected classification. QuickBird MS and PAN have been used in the study to demonstrate the enhancement and accuracy assessment of fused image by using the Segment Fusion algorithm developed and tested with their effectiveness evaluated in [10-16]. The remaining sections are organized as follows. Section 2 describes the basic terms in supervised image classification; Section 3 describes the supervised image classification methods; section 4 presents the data sets that used in the experimental analysis and classification results of fused image and Section 5 conclusions. All of the image classification speeds have been calculated, that using the same training data for each image test. The computer hardware used to record the image classification algorithm speeds are an Intel® Core ™ i5-245OM CPU@ 2.50 GHz with Turbo Boost 3.10 GHz and 4.00GB RAM installed. The ALwassaiProcess software was running on operating system Microsoft Windows 7 64-bit respectively.

## 2. BASIC TERMS IN SUPERVISED IMAGE CLASSIFICATION

In this section we would explain some basic terms about supervised image classification in general. A digital image is composed by pixels or points, and these points usually represent values in a multidimensional space. Each point can be represented as:

$$x = \begin{bmatrix} x_1 \\ x_2 \\ \vdots \\ x_M \end{bmatrix}$$

Where $x_i$ is the value of pixel x in the band or feature i (the term feature is more used since band is more related with spectral bands). The vector $x$ is also called the feature vector or measurement vector. Feature Space is the set of all possible feature vectors. Usually the value of a pixel x in a band i is the brightness or gray level for that pixel, but for some classification tasks one may want to use other features, for example, texture measures, etc. In this case it would be necessary to normalize the values in the feature space so all feature space dimensions will be the same.

Classification is a method by which unique labels are assigned to pixels based on the values of the vector $x$. This decision is made by applying a discriminant function $g_i(x)$ associated with class $\omega_i$ to vector $x$, and choosing the largest $g_i(x)$. In other words: for all classes $\omega_1, \omega_2, \dots \omega_m$ in a classification task the pixel $x$ is classified as class $\omega_i$ if its $g_i(x)$ is the largest value for all $i \in M$, or:

$$x \in \omega_i \; if \; g_i\ (x) > g_j\ (x) \; for\ all\ j \neq i \quad (1.1)$$

The total number of classes in supervised classification is determined by the nature of the problem and by user's decision. The discriminant function depends on the chosen classification method. Some discriminant functions used in the different classifiers in this ALwassaiProcess software will be presented. Classification can also be considered as the partition of the feature space in mutually exclusive parts. Pixels are assigned to classes based on this partition.

Here is a little more formal definition of the above, which is known as Bayesian Classifier, The classes in a classification task can be denoted by:

$$\omega_i, \quad i = 1, \dots.. M$$

Where $M$ is the total number of classes and the probability that the correct class for $x$ is $\omega_i$ is given by:

$$p(\omega_i\ |x), \qquad i = 1, \dots.. M$$

Where $p(\omega_i\ |x)$ is called the a-posteriori probability. To decide which class $i$ is the best (or has the least classification error) for the pixel $x$ we should select the largest $p(\omega_i\ |x)$ on other words, select from between the probabilities that the correct class for pixel $x$ is $\omega_i$ for all $i$ the highest one, or [17]:

$$x \in \omega_i \; if \; p(x|\omega_i) > p(x|\omega_j) \; for\ all\ j \neq i \quad (1.2)$$

The problem is that these $p(x|\omega_i)$ we need to determine the class for pixel $x$ are unknown. If we have enough samples from all classes we can estimate the probability for finding a pixel from class $\omega_i$ in position $x$, denoted by $p(x|\omega_i)$ . If there are $M$ classes, there will be also $M$ values for $p(x|\omega_i)$ denoting the relative probabilities that the pixel x belongs to the class $i$. The relation between $p(\omega_i|x)$ and $p(x|\omega_i)$ is given by Bayes' theorem [17]:

$$p(\omega_i\ |x) = p(x|\omega_i)\, p(\omega_i)/p(x)$$

Where $p(\omega_i)$ is the probability that class $\omega_i$ occurs in the image. Also called a-priori probability and $p(x)$ is the probability of finding a pixel from any class at position $x$. By comparison the $p(\omega_i\ |x\ )$ are posterior probabilities. Using (1.2) it can be seen that the classification rule of (1.1) is [17]:

$$x \in \omega_i \; if \; p(x|\omega_i)p(\omega_i) > p(x|\omega_j)p(\omega_j) \; for\ all\ j \neq i \quad (1.3)$$

And substituting

$$g_i\ (x) = ln\{p(x|\omega_i)p(\omega_i\ )\}$$
$$= ln\, p(x|\omega_i\ ) + ln\, p(\omega_i\ ) \qquad (1.4)$$

Where $ln$ is the natural logarithm, so that (1.3) is restated as [17]:

$$x \in \omega_i \; if \; g_i\ (x) > g_j\ (x) \; for\ all\ j \neq i \qquad (1.5)$$

Where $g_i(x)$ the discriminant function and is calculated differently for the different classification schemes. When $p(\omega_i)$ is not available it is considered as equal to 1 for all classes.

## 3. SUPERVISED IMAGE CLASSIFICATION METHODS

This study implied different discriminant functions considered as the classification strategy and will be used to classifier the training set from each of the defined $i$ classes as below. In the following different discriminant functions the training data will be extracted by having certain regions and they will have their RGB values represented by the mean red, the mean blue and the mean green values separately. Supposing the mean vector is a n-dimension vector, where n is the number of features (or image bands).

The mean vector $\mu_i$ for class $i$ is calculated for each feature as:

$$\mu_i = \frac{1}{N}\sum_{j=1}^{N} x_j$$

Where $N$ is the number of samples and $x_j$ is the jth sample

## 1) Parallelepiped Classifier (Pp)

The Pp classifier is a very simple supervised classifier, also known as a box or Level-Slice classifier. The Pp classifier method is implemented by defining a Pp classifier -like subspace (i.e., hyper-rectangle) for each class. The boundaries of the Pp classifier, for each feature, can be defined by the minimum and maximum pixel values in the given class, or alternatively, by a certain number of standard deviations on either side of the mean of the training data for the given class [18, 19]. The classification is done by checking whether the pixel is inside or outside the bounds for each feature space lies inside any of the parallelepipeds. An example illustrating the specification of the topology of a Pp classifier in the case of a two-dimensional feature space is shown in Fig. 1.

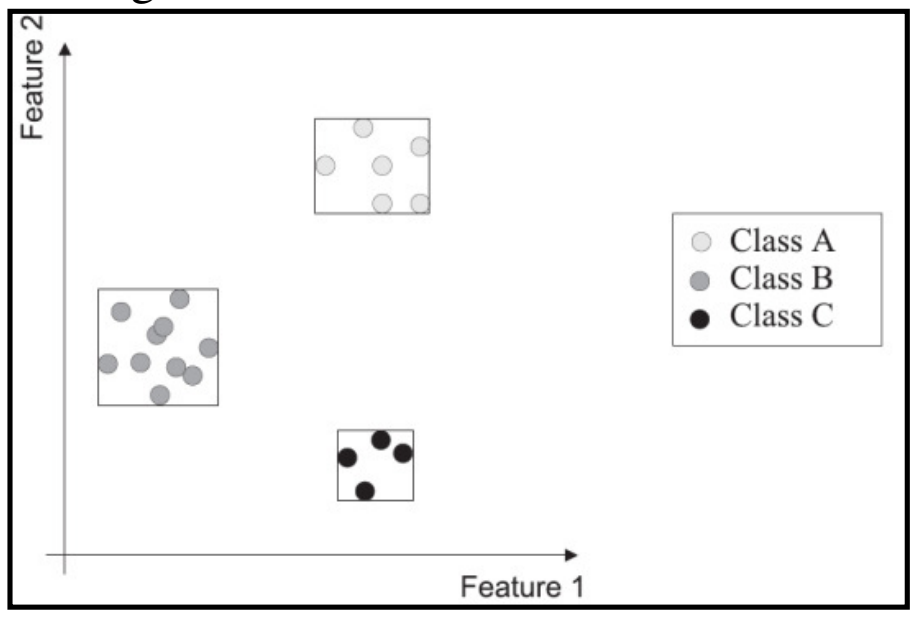

Fig. 1: Pp Classifier Example

## 2) Mahalanobis Distance Classifier (MD)

Classification is performed on MD classifier from each pixel to the signatures centers. Basically the classifier assigns class $i$ to pixel $x$ if:

$$g_i\ (x) < g_j\ (x) \quad for\ all\ \ i \neq j$$

The discriminant function for the Mahalanobis distance classifier is as follows [19-20]:

$$g_i\ (x)=(x-\mu_i)^t {\Sigma_i}^{-1}(x-\mu_i) \qquad (1.6)$$

Where $g_i\ (x)$ is the MD classifier for class $i$. $\mu_i$ and ${\Sigma_i}^{-1}$ are the mean vector and the inverse covariance matrix for the data of class $i$. For a MD classifier signature we need some of the components shown in equation 1.6: the classes' mean and inverse of the covariance matrix ${\Sigma_i}^{-1}$. The data will be stored in separate planes in the depth direction, one for the mean and one for the inverse of the covariance matrix. The classification is done by choosing the lowest $g(x)$ for all class $\in N$ .

## 3) Maximum-Likelihood Classifier (ML)

The ML classifier assumes that the classes are unimodal and normally distributed. Its discriminant function is given by [19]:

$$g_i(x) = ln\,p(\omega_i) - \frac{1}{2}\, ln\,|\Sigma_i|\ (x - \mu_i)^t {\Sigma_i}^{-1} (x - \mu_i)\} \qquad (1.7)$$

Often the analyst has no useful information about the $p(\omega_i)$in which case a situation of equal prior probabilities is assumed; as a result $\ln p(\omega_i)$ can be removed from (1.7) since it is then the same for all $i$. In that case the 1/2 common factor can also be removed leaving, as the discriminant function [19]:

$$g_i(x) = -ln|\Sigma_i| - (x - \mu_i)^t\, {\Sigma_i}^{-1}\, (x - \mu_i) \quad (1.8)$$

Implementation of the ML decision rule involves using either (1.7) or (1.8) in (1.5). For a ML classifier signature we need some of the components shown in equation 1.8: the classes' mean vector $\mu_i$ and inverse of the covariance matrix ${\Sigma_i}^{-1}$ . To help with the analysis of the signature's distributions and speed calculation of the likelihoods, the covariance matrix $\Sigma_i$ and the negative of the logarithm of its determinate will also be stored in the signature. The data will be stored in separate planes in the depth direction, one for the mean, one for the covariance matrix, one for the inverse of the covariance matrix and one for the negative logarithm of the determinate of the covariance matrix. The classification is done by choosing the maximum $g(x)$ for all class $i \in N$ . $N$ is the number of features in the image.

## 4) Euclidean Distance Classifier (ED)

The ED is a particular case of Minkowski sometimes is also called Quadratic Mean. Classification is performed on ED classifier from each pixel to the signatures centers. Basically the classifier assigns class $i$ to pixel $x$ if:

$$g_i\ (x) < g_j\ (x) \quad for\ all\ \ i \neq j$$

The discriminant function for the EC classifier takes the following form and has the unit circle detailed in [21]:

$$g_i(x) = \left(\sum_i^n |x - \mu_i|^2\right)^{1/2} = \sqrt{|x_1 - \mu_1|^2 + \ldots + |x_n - \mu_n|^2} \qquad (1.9)$$

# 4. EXPERIMENTAL RESULTS

## 4.1 Test data sets

The images that are going to be fused and classified in this study are downloaded from http://studio.gge.unb.ca/UNB/images. These remote sensing images are taken by QuickBird satellite sensor which collects one panchromatic band (450-900 nm) of the 0.7 m resolution and blue (450-520 nm), green (520-600 nm), red (630-690 nm), near infrared (760-900 nm) bands of the 2.8 m resolution. The coverage of the images was over the Pyramid area of Egypt in 2002 as shown in Fig.2. Before the image fusion, the raw MS were resampled to the same spatial resolution of the PAN in order to perform image registration. The test images of size 864 by 580 at the resolution of 0.7 m are cut from the raw images. The classification is tested on resulted image fused by using the SF algorithm as shown in Fig.2.

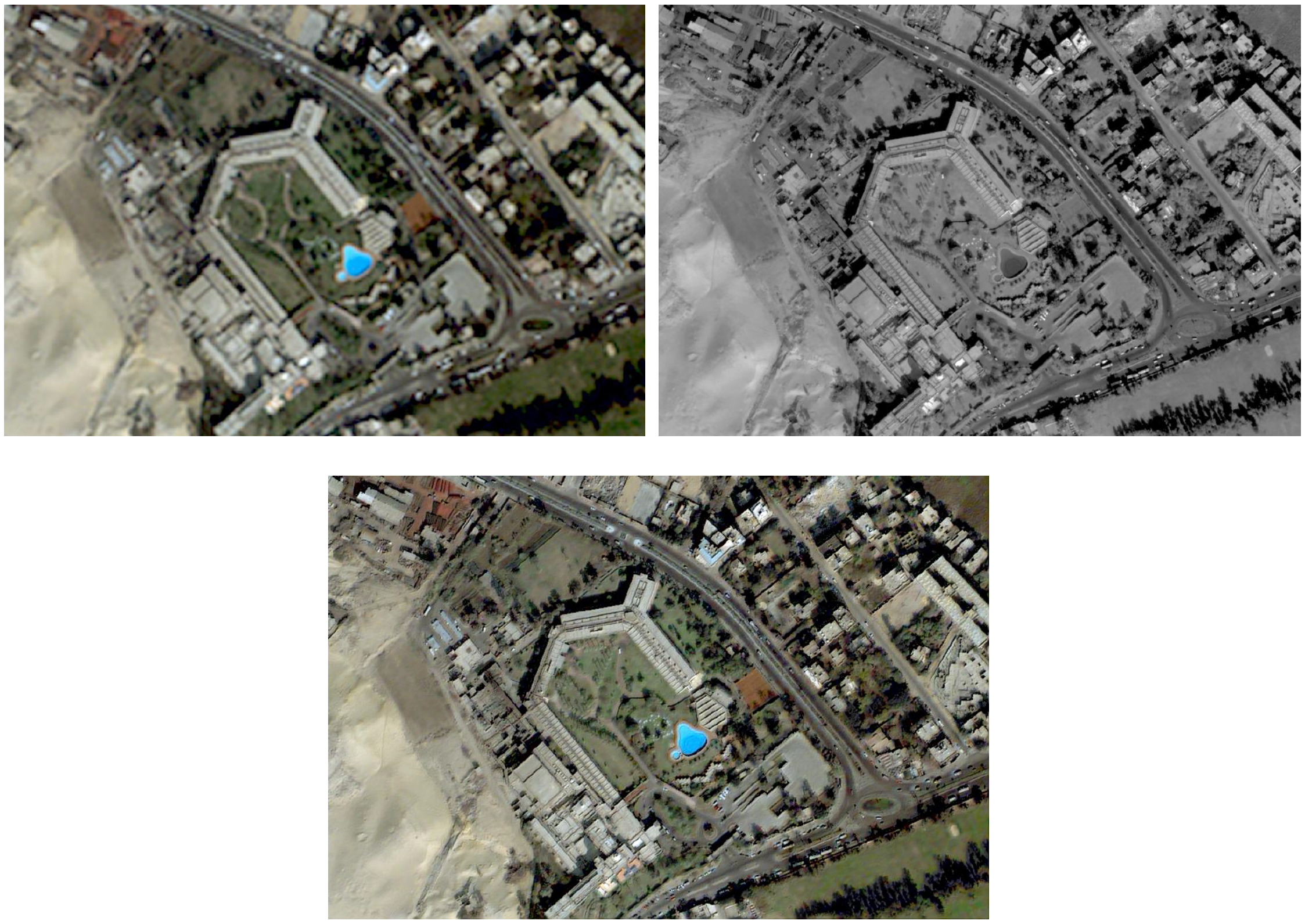

Fig. 2: Experimental Test Images over the Pyramid Area of Egypt in 2002. Quickbird Data MS and Quickbird PAN, Image Fused Resulted By SF Algorithm.

### 4.2 Supervised Image Classifiers

In the supervised classification, the acquisition of ground truth data for training and assessment is a critical component in process. In the study the developed system was designed to extract the training data test by having certain regions selected as decried below. The classification consists of the following steps:

- Step 1: Select the number and the size of regions that will be the training data the image as shown in Fig.3. The author has selected twelve classes as shown in Fig.4, and the size of each region selecting for the training data is 4 × 4 pixels was chosen.
- Step 2: Experts the Image; Experts training data; and Select discriminant function as shown in Fig.5.
- Step 3: apply the decision rule between the pixels in the image and every reference class *i* according to the selected discriminant function as shown in Fig.6.
- Step 4: Assign each pixel x to the reference class i that has the decision rule between pixel x and reference class i then stored in separate planes in the depth direction.
- Step 5: selected different five regions of each reference class *i* for the accuracy assessment of image classification as shown in Fig.7
- Step 5: The Accuracy Assessment of Image Classification as shown in Fig.8.

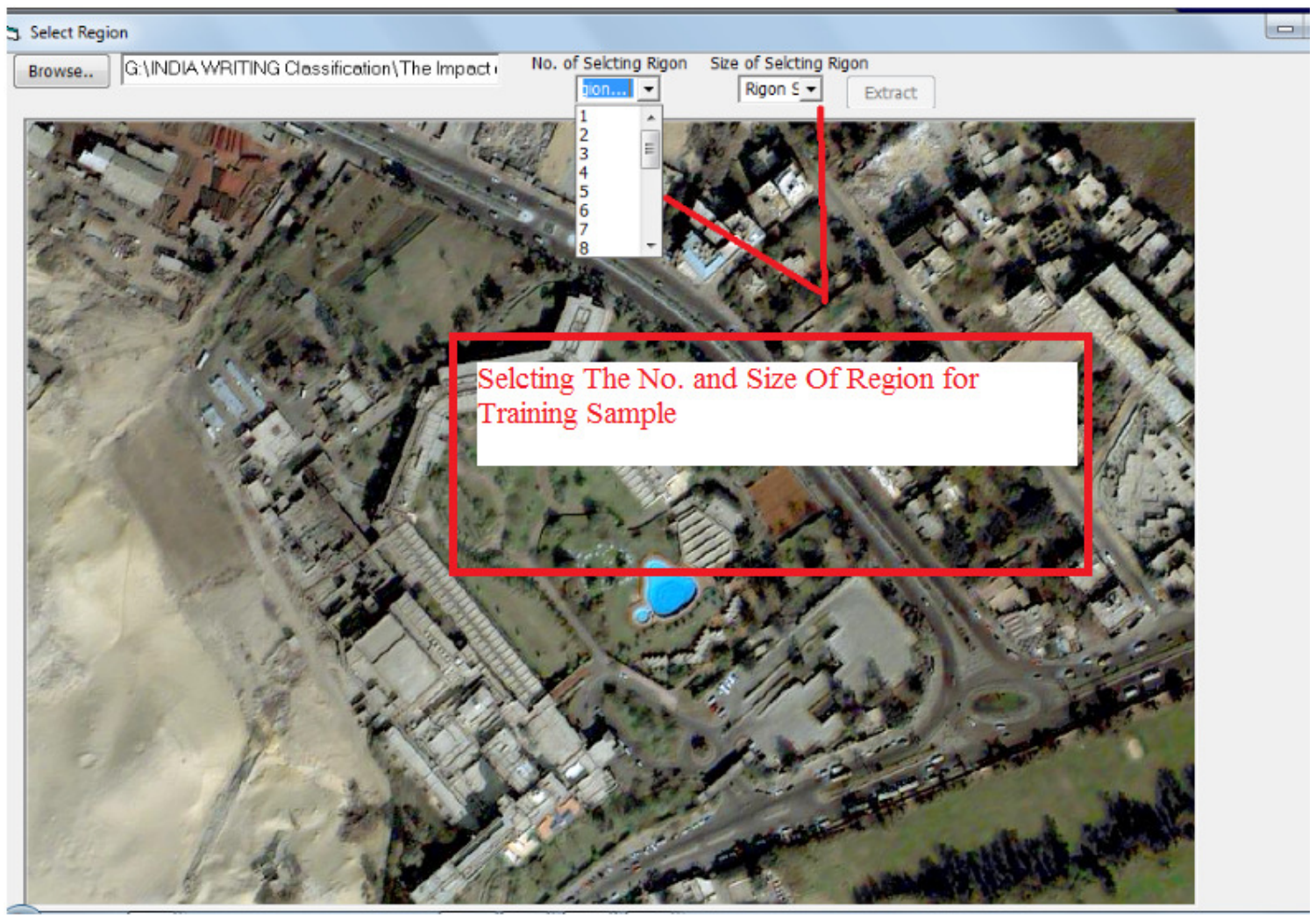

Fig.3: Illustrate Step 1: Select The Number And The Size Of Regions for The Training Data The Image

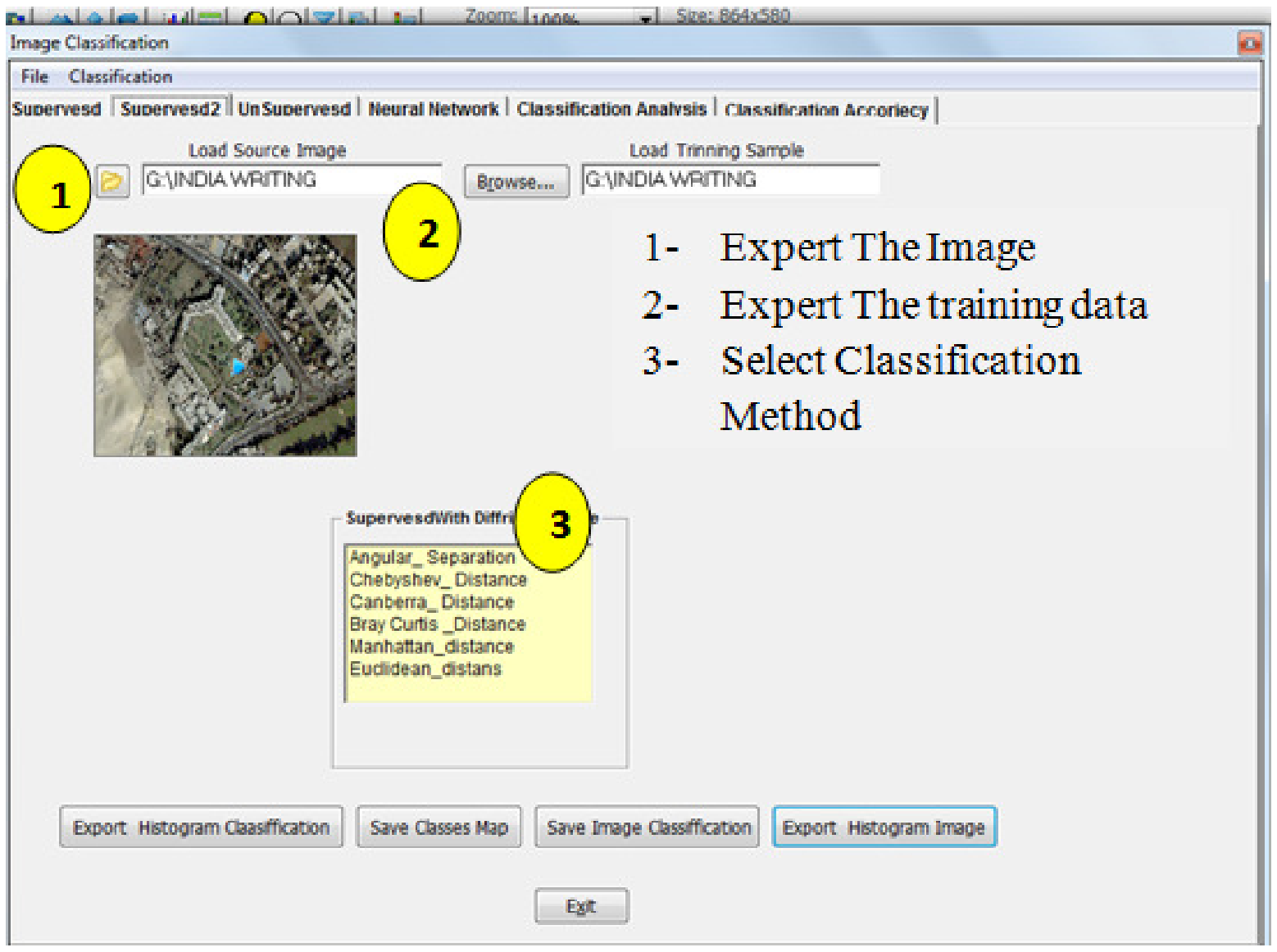

Fig.5: Illustrate Step 2: the Automatic Classification Process: Experts The Image; Experts Training Data; And Select Classifier Methods.

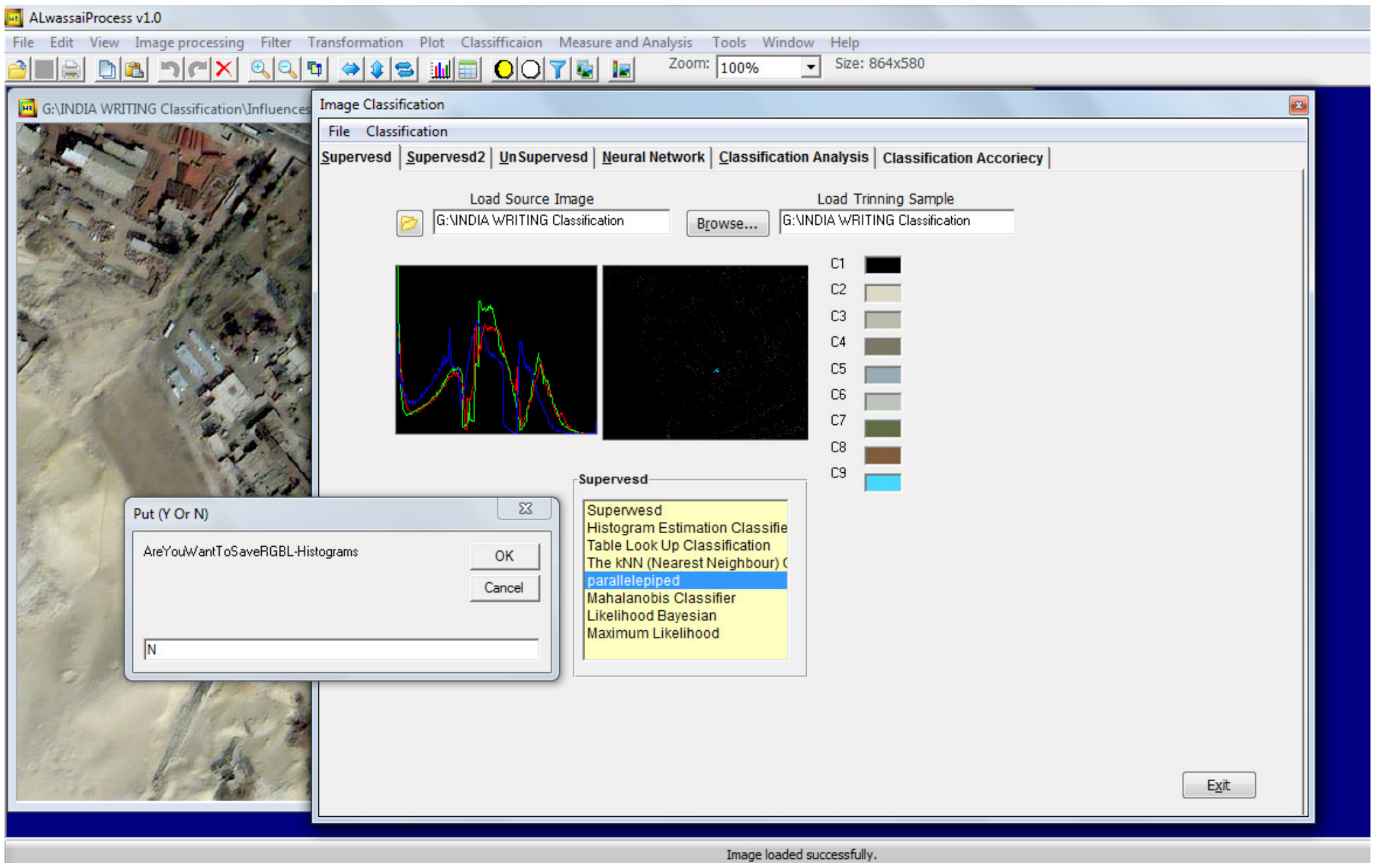

Fig.6: Illustrate Step 3: Apply The Decision Rule Between The Pixels In The Image And Every Reference Class i According To The Selected Discriminant Function.

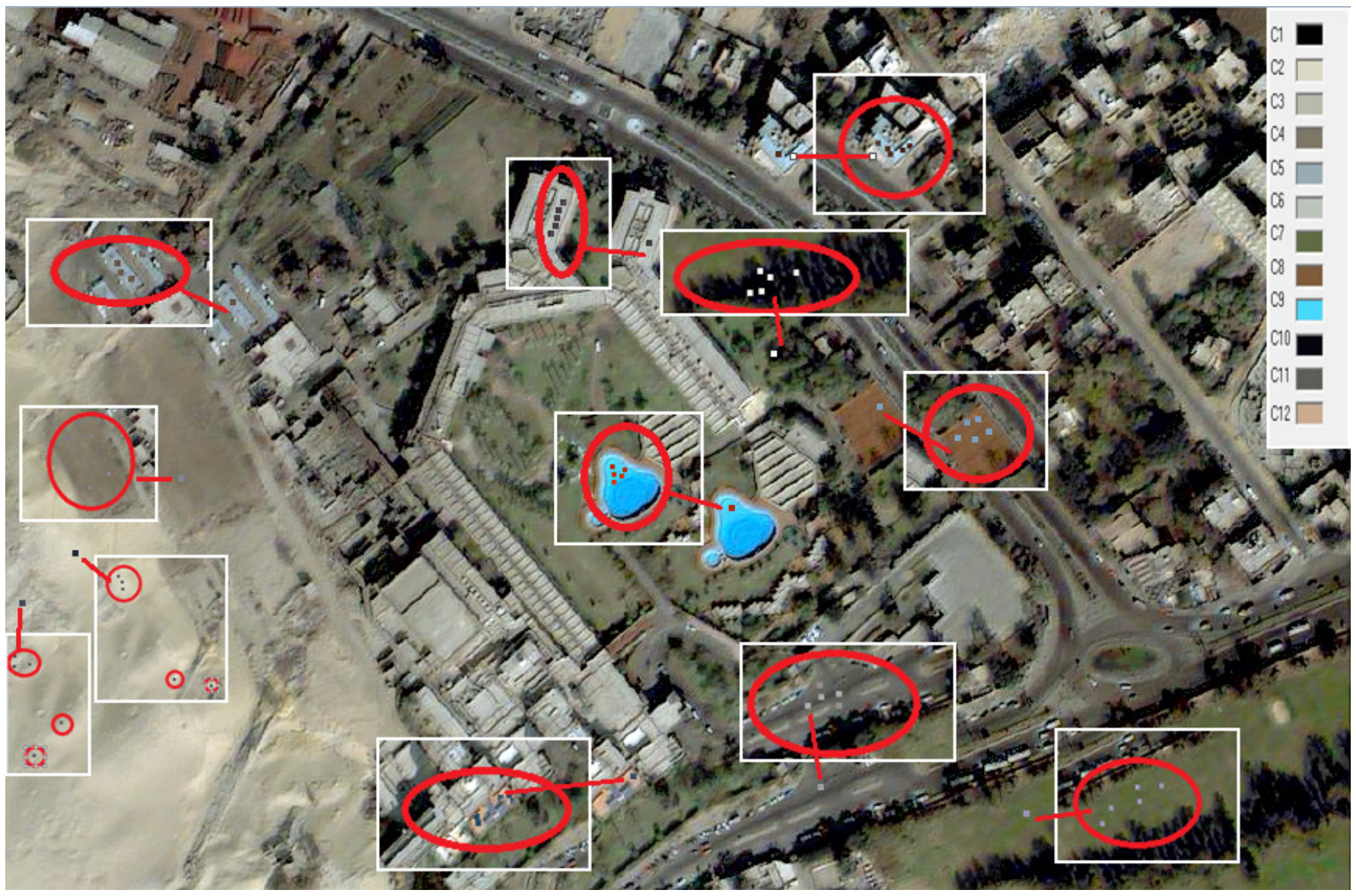

Fig.7: Illustrate Step 4: The Selected Different Five Regions Of Each Reference Class For Accuracy Assessment Of Image Classification

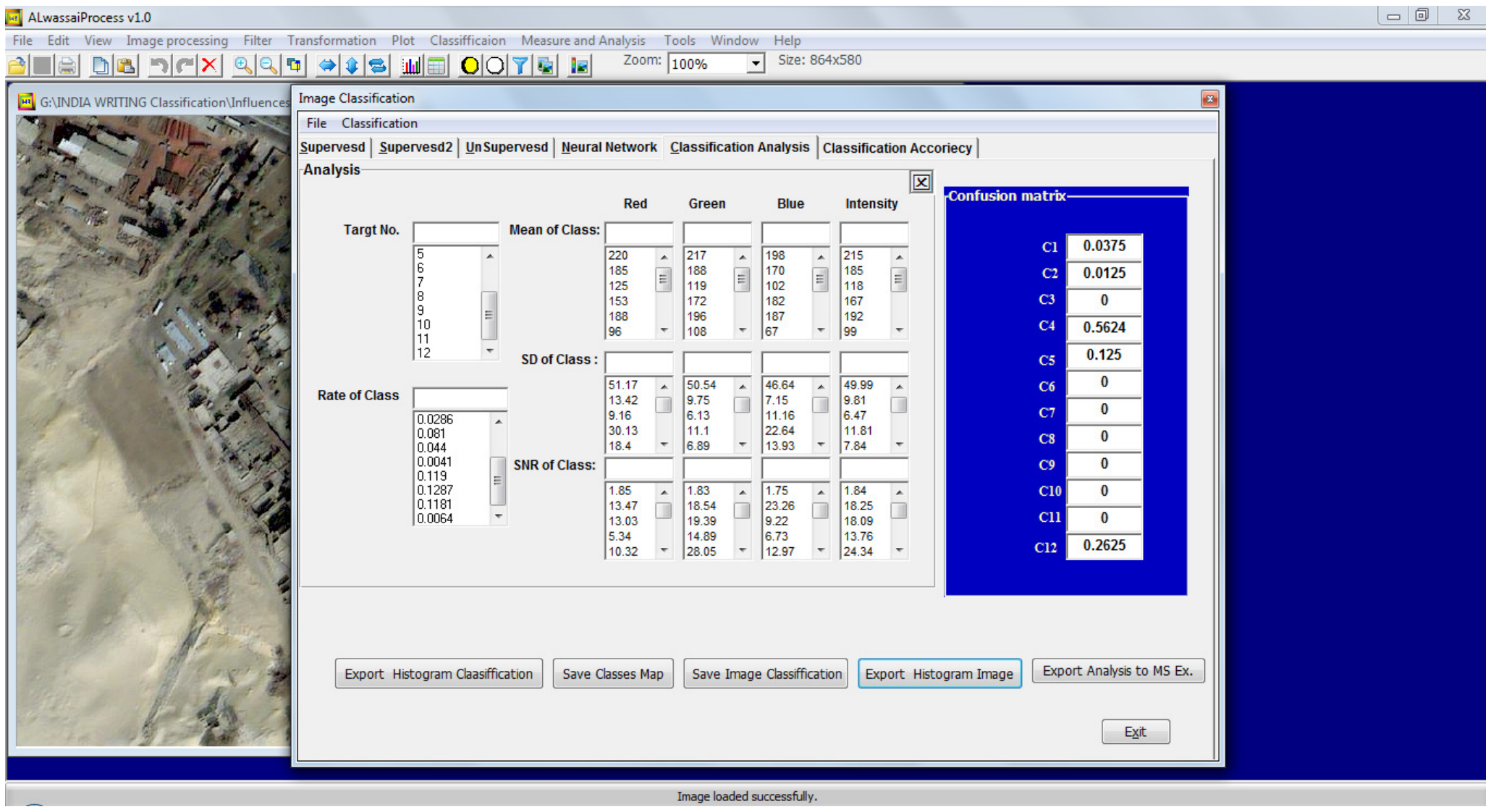

Fig8: Illustrate Step 5: The Accuracy Assessment Of Image Classification.

### 4.3 Classification Results Of Fused Image

To evaluate the performance of the proposed active learning strategies the four classifiers were applied for both MS QuickBird and fusion data after the fusion process. To the description of classification error, it is necessary to configure the error matrix and decide the measurements. Generally, there are descriptive statistic and analytic statistic from the error matrix. Overall accuracy, producer's accuracy (omission error) and user's accuracy (commission error) as well as Kappa statistic belong to descriptive statistic. In this study, as limited time, we focus the accuracy assessment of image classification only on the Overall accuracy for fused image. For such purpose, we first selected different five regions that have a 4x4 size for each reference class set is shown above in Fig.4. Table (1- 4) and Table (5-8) list the error matrix for both classified results, respectively. The overall measured accuracies of the Pp, MD, ML and ED classifiers for MS were 59.735%; 64.60%; 59.108% and 87.257% respectively, and for fused image classified results were 63.498%; 60.838%; 58.363 % and 91.476% respectively. Fig. 9 show the classified results for fusion image and MS QuickBird image by the four classifiers

The Pp classifier is quick and easy to implement but the classification results has error, because some pixels lie inside more than one Pp or outside all Pps, therefore a pixel in those regions not classified. The classification results of the ML and MD classifiers are not surprising when we consider the fact that both ML and MD classifiers use class variances in each spectral band for calculating distances for classification. Both MD and ML classifiers use parametric rules that require normally distributed data and well defined variances for image data and each training class, while most image data do not show normal distribution, and most of the training classes have high variances of pixel values in each band. The ML and MD classifiers rules can be quite diagnostic in distinguishing different features with image data that show normal distribution and have well defined variances in each spectral band for each surface object. However, when those assumptions are violated, their performances are less than desirable. Classification accuracy results of the supervised ED classifier are presented in Table 8 with the overall accuracy of 91.476percent. The mapping result of the ED classifier shows much higher overall accuracy of 91.476 % compared to that of the ML classifier (58.363 %) or MD classifier (60.838%) and Pp classifier (63.498%). In general, the supervised classification results of fusion image generated better than did the MS QuickBird. The best results overall accuracies with ED classifier than the other did.

Table (1): Error Matrix for MS QuickBird Classified Result By Pp Classifier

| | C1 | C2 | C3 | C4 | C5 | C6 | C7 | C8 | C9 | C10 | C11 | C12 | R. Total |
|---|---|---|---|---|---|---|---|---|---|---|---|---|---|
| **C1** | **0.6999** | 0.1 | | 0.0625 | 0.1375 | | | | | | | | 0.9999 |
| **C2** | | **0.4499** | | 0.1625 | 0.3874 | | | | | | | | 0.9998 |
| **C3** | 0.3624 | | **0.5124** | | | | | | | 0.125 | | | 0.9998 |
| **C4** | 0.1093 | 0.2343 | | **0.6093** | | | 0.0156 | | | | 0.0312 | | 0.9997 |
| **C5** | 0.0375 | | | | **0.9624** | | | | | | | | 0.9999 |
| **C6** | 0.2874 | | | | | **0.6124** | 0.075 | | 0.025 | | | | 0.9998 |
| **C7** | | | 0.0125 | | | 0.0125 | **0.9749** | | | | | | 0.9999 |
| **C8** | 0.0222 | | | 0.0444 | 0.2222 | 0.0888 | | **0.6222** | | | | | 0.9998 |
| **C9** | | | | | | | | | **0.9999** | | | | 0.9999 |
| **C10** | 0.3249 | | 0.075 | | | 0.125 | 0.3499 | | | **0.125** | | | 0.9998 |
| **C11** | 0.2 | 0.0125 | 0.25 | | 0.2 | | | | | | **0.3374** | | 0.9999 |
| **C12** | 0.0375 | 0.0125 | | 0.5624 | 0.125 | | | | | | | **0.2625** | 0.9999 |
| **C.Total** | 2.0811 | 0.8092 | 0.8499 | 1.4411 | 2.0345 | 0.8387 | 1.4154 | 0.6222 | 1.0249 | 0.25 | 0.3686 | 0.2625 | 11.9981 |
| **Overall accuracy** | 0.6999 | 0.4499 | 0.5124 | 0.6093 | 0.9624 | 0.6124 | 0.9749 | 0.6222 | 0.9999 | 0.125 | 0.3374 | 0.2625 | **0.59735** |

Table (2): Error Matrix for MS QuickBird Classified Result By MD Classifier

| | C1 | C2 | C3 | C4 | C5 | C6 | C7 | C8 | C9 | C10 | C11 | C12 | R. Total |
|---|---|---|---|---|---|---|---|---|---|---|---|---|---|
| **C1** | **0.6874** | | | | 0.3124 | | | | | | | | 0.9998 |
| **C2** | | **0.1** | | | 0.8999 | | | | | | | | 0.9999 |
| **C3** | | | **0.7624** | | 0.15 | | 0.0875 | | | | | | 0.9999 |
| **C4** | | | | **0.3906** | 0.5937 | | | | | | 0.0156 | | 0.9999 |
| **C5** | | | | | **0.9999** | | | | | | | | 0.9999 |
| **C6** | | | | | | **0.9999** | | | | | | | 0.9999 |
| **C7** | | | | | | 0.0125 | **0.9874** | | | | | | 0.9999 |
| **C8** | | | | | 0.4 | | | **0.6** | | | | | 1 |
| **C9** | | | | | | | | | **0.9999** | | | | 0.9999 |
| **C10** | | | 0.2125 | | | 0.1125 | 0.05 | | | **0.6249** | | | 0.9999 |
| **C11** | | | | | 0.5999 | | | | | | **0.3999** | | 0.9998 |
| **C12** | | | | 0.025 | 0.7749 | | | | | | | **0.2** | 0.9999 |
| **C.Total** | 0.6874 | 0.1 | 0.9749 | 0.4156 | 4.7307 | 1.1249 | 1.1249 | 0.6 | 0.9999 | 0.6249 | 0.4155 | 0.2 | 11.9987 |
| **Overall accuracy** | 0.6874 | 0.1 | 0.7624 | 0.3906 | 0.9999 | 0.9999 | 0.9874 | 0.6 | 0.9999 | 0.6249 | 0.3999 | 0.2 | **0.646025** |

Table (3): Error Matrix for MS QuickBird Classified Result By ML Classifier

| | C1 | C2 | C3 | C4 | C5 | C6 | C7 | C8 | C9 | C10 | C11 | C12 | R. Total |
|---|---|---|---|---|---|---|---|---|---|---|---|---|---|
| **C1** | **0.15** | | | | 0.8499 | | | | | | | | 0.9999 |
| **C2** | | **0.0125** | | | 0.9874 | | | | | | | | 0.9999 |
| **C3** | | | **0.7499** | | 0.15 | | 0.1 | | | | | | 0.9999 |
| **C4** | | | | **0.3437** | 0.5 | | | | | | 0.1562 | | 0.9999 |
| **C5** | | | | | **0.9999** | | | | | | | | 0.9999 |
| **C6** | | | | | | **0.9999** | | | | | | | 0.9999 |
| **C7** | | | | | | | **0.9999** | | | | | | 0.9999 |
| **C8** | | | | 0.4 | | | | **0.6** | | | | | 1 |
| **C9** | | | | | | | | | **0.9999** | | | | 0.9999 |
| **C10** | | | 0.1875 | | | 0.1 | 0.075 | | | **0.6374** | | | 0.9999 |
| **C11** | | | | | 0.5999 | | | | | | **0.3999** | | 0.9998 |
| **C12** | | | | 0.025 | 0.7749 | | | | | | | **0.2** | 0.9999 |
| **C.Total** | 0.15 | 0.0125 | 0.9374 | 0.7687 | 4.862 | 1.0999 | 1.1749 | 0.6 | 0.9999 | 0.6374 | 0.5561 | 0.2 | 11.9988 |
| **Overall accuracy** | 0.15 | 0.0125 | 0.7499 | 0.3437 | 0.9999 | 0.9999 | 0.9999 | 0.6 | 0.9999 | 0.6374 | 0.3999 | 0.2 | **0.591083** |

Table (4): Error Matrix for MS QuickBird Classified Result By ED Classifier

| Euclidian | C1 | C2 | C3 | C4 | C5 | C6 | C7 | C8 | C9 | C10 | C11 | C12 | R. Total |
|---|---|---|---|---|---|---|---|---|---|---|---|---|---|
| **C1** | **0.9749** | | | | 0.025 | | | | | | | | 0.9999 |
| **C2** | 0.0375 | **0.7874** | | | 0.1625 | | | | | | 0.0125 | | 0.9999 |
| **C3** | | | **0.9999** | | | | | | | | | | 0.9999 |
| **C4** | | | 0.0781 | **0.9218** | | | | | | | | | 0.9999 |
| **C5** | | 0.25 | | 0.05 | **0.6999** | | | | | | | | 0.9999 |
| **C6** | | | | | | **0.8999** | | | | | 0.1 | | 0.9999 |
| **C7** | | | | | | | **0.9999** | | | | | | 0.9999 |
| **C8** | | | | | | | | **1** | | | | | 1 |
| **C9** | | | | | | | | | **0.9624** | 0.0375 | | | 0.9999 |
| **C10** | | | 0.2874 | | | | | | | **0.7124** | | | 0.9998 |
| **C11** | 0.0375 | 0.0375 | 0.2125 | | | | | | | 0.025 | **0.6874** | | 0.9999 |
| **C12** | 0.0125 | | 0.0375 | 0.0875 | 0.0375 | | | | | | | **0.8249** | 0.9999 |
| **Total** | 1.0624 | 1.0749 | 1.6154 | 1.0593 | 0.9249 | 0.8999 | 0.9999 | 1 | 0.9624 | 0.7749 | 0.7999 | 0.8249 | 11.9988 |
| **overall accuracy** | 0.9749 | 0.7874 | 0.9999 | 0.9218 | 0.6999 | 0.8999 | 0.9999 | 1 | 0.9624 | 0.7124 | 0.6874 | 0.8249 | **0.872566667** |

Table (5): Error Matrix Classified Result for Fusion Image By Pp Classifier

| | **C1** | **C2** | **C3** | **C4** | **C5** | **C6** | **C7** | **C8** | **C9** | **C10** | **C11** | **C12** | **R. Total** |
|---|---|---|---|---|---|---|---|---|---|---|---|---|---|
| **C1** | **0.9749** | 0.025 | | | | | | | | | | | 0.9999 |
| **C2** | 0.025 | **0.5249** | | 0.0625 | 0.3249 | | | | | | 0.0625 | | 0.9998 |
| **C3** | 0.125 | | **0.8624** | | | | 0.0125 | | | | | | 0.9999 |
| **C4** | | 0.5 | 0.0312 | **0.3125** | 0.0312 | | 0.0937 | | | | 0.0312 | | 0.9998 |
| **C5** | | 0.3124 | | 0.0125 | **0.6499** | | | | | | 0.025 | | 0.9998 |
| **C6** | | | 0.0375 | | | **0.8124** | 0.025 | | | 0.125 | | | 0.9999 |
| **C7** | | | 0.0125 | | | 0.125 | **0.8624** | | | | | | 0.9999 |
| **C8** | 0.1555 | | | | 0.2444 | | | **0.1333** | 0.0222 | | | 0.4444 | 0.9998 |
| **C9** | | | | | | | | | **0.9999** | | | | 0.9999 |
| **C10** | 0.2874 | | 0.2375 | | | 0.1625 | | | | **0.3124** | | | 0.9998 |
| **C11** | 0.1625 | 0.075 | 0.1125 | | 0.0375 | 0.0125 | | | | 0.0125 | **0.5499** | 0.0375 | 0.9999 |
| **C12** | 0.05 | 0.025 | 0.0375 | 0.0125 | 0.175 | | | | | | 0.075 | **0.6249** | 0.9999 |
| **C.Total** | 1.7803 | 1.4623 | 1.3311 | 0.4 | 1.4629 | 1.1124 | 0.9936 | 0.1333 | 1.0221 | 0.4499 | 0.7436 | 1.1068 | 11.9983 |
| **Overall accuracy** | 0.9749 | 0.5249 | 0.8624 | 0.3125 | 0.6499 | 0.8124 | 0.8624 | 0.1333 | 0.9999 | 0.3124 | 0.5499 | 0.6249 | **0.634983** |

Table (6): Error Matrix Classified Result for Fusion Image By MD classifier

| | **C1** | **C2** | **C3** | **C4** | **C5** | **C6** | **C7** | **C8** | **C9** | **C10** | **C11** | **C12** | **R. Total** |
|---|---|---|---|---|---|---|---|---|---|---|---|---|---|
| **C1** | **0.5874** | | | | 0.3374 | | | | | | | 0.075 | 0.9998 |
| **C2** | | **0.0875** | | | 0.4749 | | | | | | 0.4374 | | 0.9998 |
| **C3** | | | **0.4749** | | | | | | | 0.3624 | 0.1625 | | 0.9998 |
| **C4** | | | | **0.25** | 0.0156 | | | | | 0.0625 | 0.2031 | 0.4687 | 0.9999 |
| **C5** | | | | | **0.9624** | | | | | | 0.0125 | 0.025 | 0.9999 |
| **C6** | | | | | | **0.15** | | | | 0.8499 | | | 0.9999 |
| **C7** | | | | | | 0.2625 | **0.6249** | | | 0.1125 | | | 0.9999 |
| **C8** | | | | | | | | **0.2888** | | | | 0.7111 | 0.9999 |
| **C9** | | | | | | | | | **0.9999** | | | | 0.9999 |
| **C10** | | | | | | | | | | **0.9999** | | | 0.9999 |
| **C11** | | | | | 0.0125 | | | | | 0.0625 | **0.9249** | | 0.9999 |
| **C12** | | | | 0.0125 | | | | | | 0.0125 | 0.025 | **0.9499** | 0.9999 |
| **C. Total** | 0.5874 | 0.0875 | 0.4749 | 0.2625 | 1.8028 | 0.4125 | 0.6249 | 0.2888 | 0.9999 | 2.4622 | 1.7654 | 2.2297 | 11.9985 |
| **Overall accuracy** | 0.5874 | 0.0875 | 0.4749 | 0.25 | 0.9624 | 0.15 | 0.6249 | 0.2888 | 0.9999 | 0.9999 | 0.9249 | 0.9499 | **0.608375** |

Table (7): Error Matrix Classified Result for Fusion Image By ML Classifier

| | **C1** | **C2** | **C3** | **C4** | **C5** | **C6** | **C7** | **C8** | **C9** | **C10** | **C11** | **C12** | **R. Total** |
|---|---|---|---|---|---|---|---|---|---|---|---|---|---|
| **C1** | **0.3999** | | | | 0.4374 | | | | | | | 0.1625 | 0.9998 |
| **C2** | | **0.0875** | | | 0.4499 | | | | | | 0.4624 | | 0.9998 |
| **C3** | | | **0.4624** | | | | | | | 0.3624 | 0.175 | | 0.9998 |
| **C4** | | | | **0.2031** | | | | | | 0.0468 | 0.0781 | 0.6718 | 0.9998 |
| **C5** | | | | | **0.9249** | | | | | | 0.0375 | 0.0375 | 0.9999 |
| **C6** | | | | | | **0.1125** | | | | 0.8874 | | | 0.9999 |
| **C7** | | | | | | 0.2 | **0.6249** | | | 0.175 | | | 0.9999 |
| **C8** | | | | | | | | **0.2888** | | | | 0.7111 | 0.9999 |
| **C9** | | | | | | | | | **0.9999** | | | | 0.9999 |
| **C10** | | | | | | | | | | **0.9999** | | | 0.9999 |
| **C11** | | | | | | | | | | 0.0625 | **0.9374** | | 0.9999 |
| **C12** | | | | | | | | | | 0.0125 | 0.025 | **0.9624** | 0.9999 |
| **Column total** | 0.3999 | 0.0875 | 0.4624 | 0.2031 | 1.8122 | 0.3125 | 0.6249 | 0.2888 | 0.9999 | 2.5465 | 1.7154 | 2.5453 | 11.9984 |
| **Overall accuracy** | 0.3999 | 0.0875 | 0.4624 | 0.2031 | 0.9249 | 0.1125 | 0.6249 | 0.2888 | 0.9999 | 0.9999 | 0.9374 | 0.9624 | **0.583633** |

Table (8): Error Matrix Classified Result for Fusion Image By ED Classifier

| | **C1** | **C2** | **C3** | **C4** | **C5** | **C6** | **C7** | **C8** | **C9** | **C10** | **C11** | **C12** | **R. Total** |
|---|---|---|---|---|---|---|---|---|---|---|---|---|---|
| **C1** | **0.9624** | 0.025 | | | 0.0125 | | | | | | | | 0.9999 |
| **C2** | | **0.8749** | | | 0.1 | | | | | | 0.025 | | 0.9999 |
| **C3** | | | **0.9999** | | | | | | | | | | 0.9999 |
| **C4** | | | 0.1093 | **0.8906** | | | | | | | | | 0.9999 |
| **C5** | | 0.1 | | | **0.8999** | | | | | | | | 0.9999 |
| **C6** | | | | | | **0.9499** | | | | 0.05 | | | 0.9999 |
| **C7** | | | | | | | **0.9999** | | | | | | 0.9999 |
| **C8** | | | | | | | | **1** | | | | | 1 |
| **C9** | | | | | | | | | **0.9999** | | | | 0.9999 |
| **C10** | | | 0.2375 | | | | | | | **0.7624** | | | 0.9999 |
| **C11** | | | 0.1625 | | | | | | | | **0.8374** | | 0.9999 |
| **C12** | | | | 0.2 | | | | | | | | **0.7999** | 0.9999 |
| **C total** | 0.9624 | 0.9999 | 1.5092 | 1.0906 | 1.0124 | 0.9499 | 0.9999 | 1 | 0.9999 | 0.8124 | 0.8624 | 0.7999 | 11.9989 |
| **Overall accuracy** | 0.9624 | 0.8749 | 0.9999 | 0.8906 | 0.8999 | 0.9499 | 0.9999 | 1 | 0.9999 | 0.7624 | 0.8374 | 0.7999 | **0.914758** |

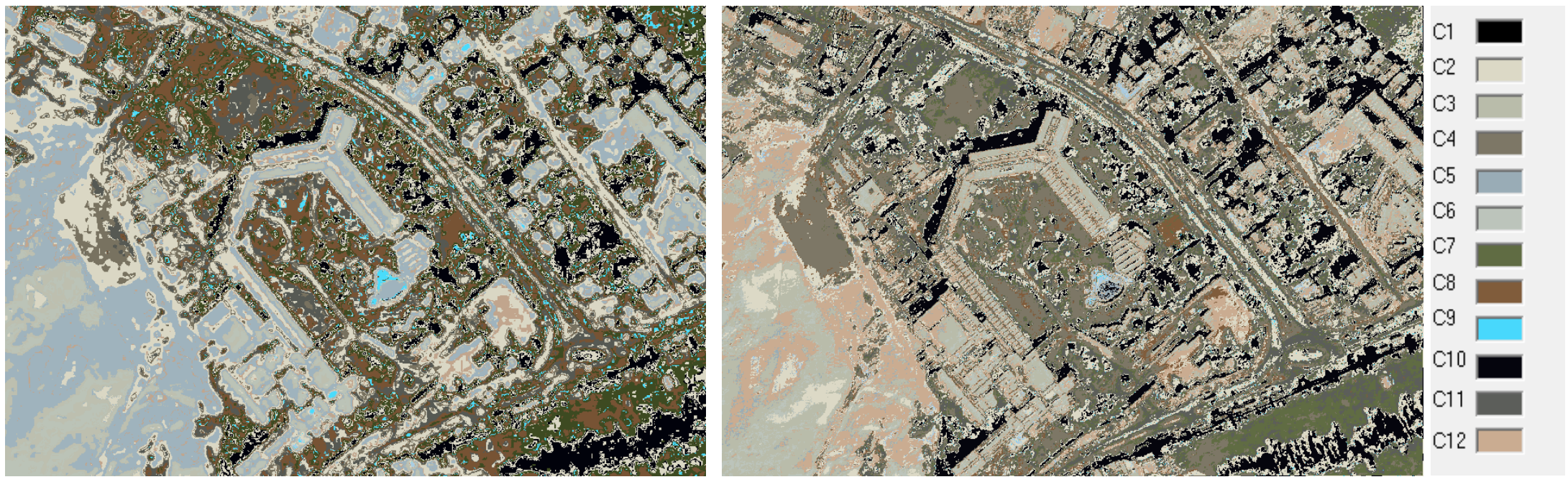

(a) PpClassifier

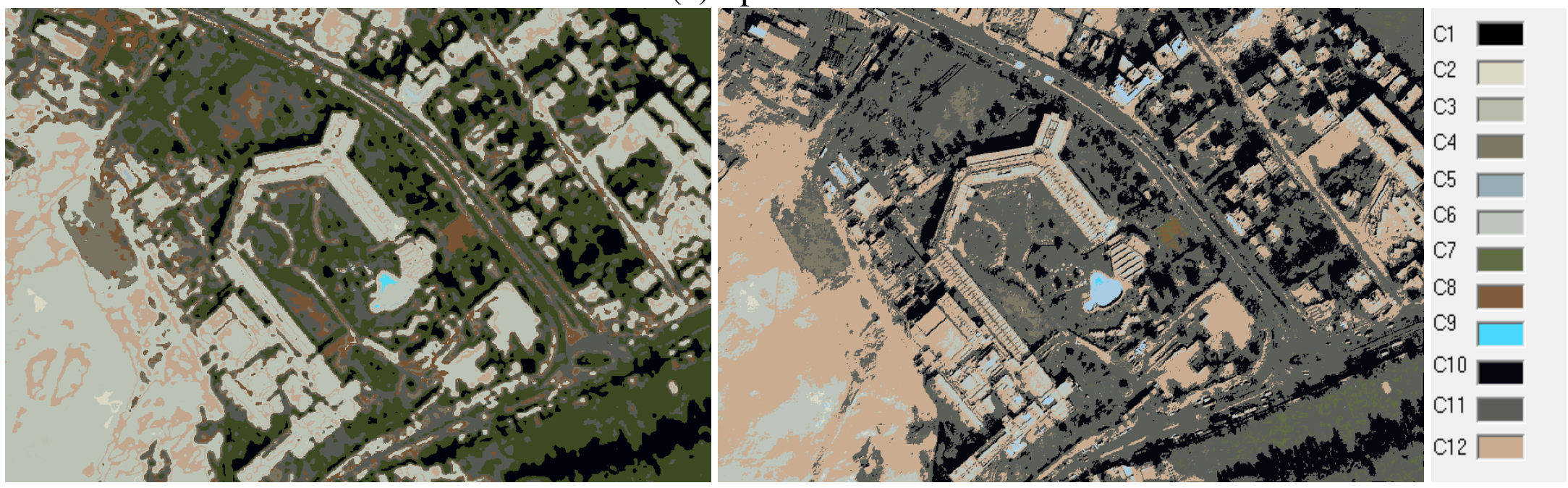

(b) MD Classifier

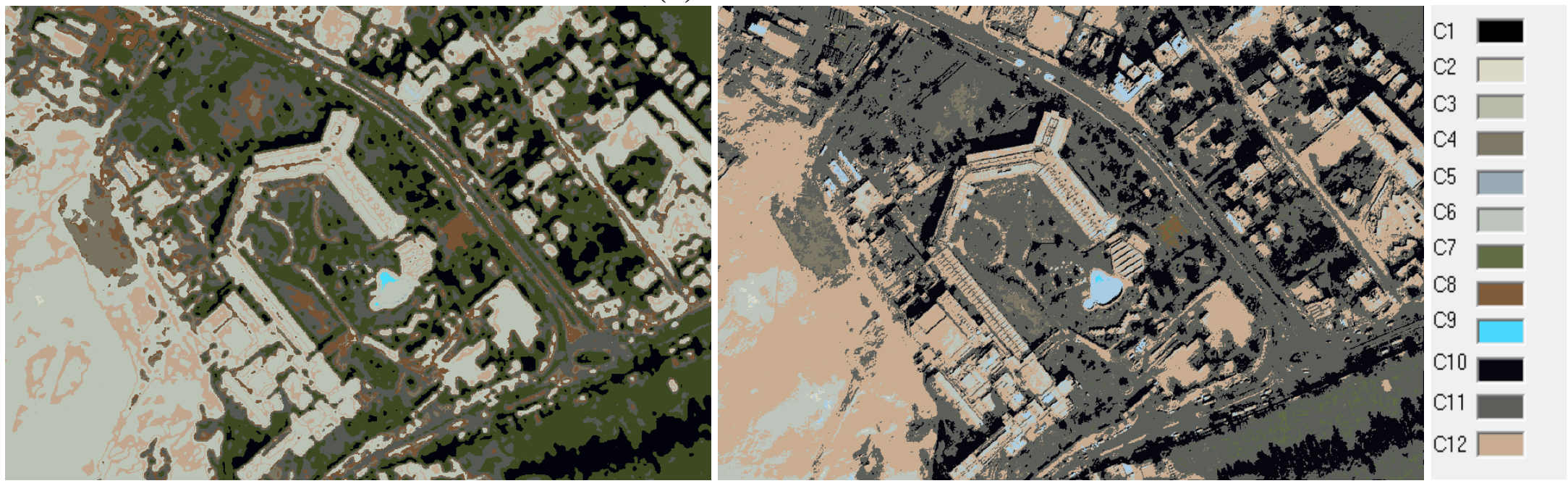

(c) ML Classifier

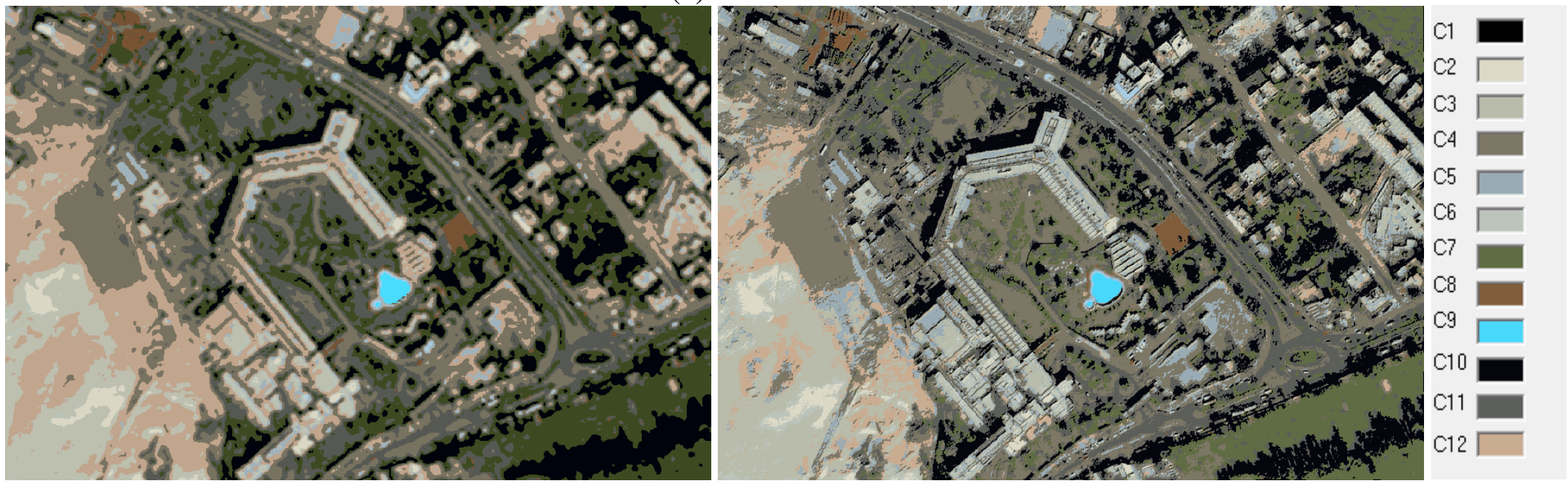

(d) ED Classifier

Fig.9: The Left Side Classified Result Of MS Quickbird And The Right Side Classified Result Of Fusion Image With Color Code Of Each Land Class By: (a) Pp Classifier (b) MD classifier; (c) ML Classifier; (d) ED Classifier.

## 5. Conclusion

In the study there are four supervised classifications are introduced as the following: Pp, MD, ML and ED classifiers. The supervised classification of the MS QuickBird Classified image has the lowest accuracy in comparison of the Fused Image Classified Result. When two data sets combined together (MS and PAN images) by using the SF algorithm in feature-level image fusion, confusion, problem was solved effectively. Another advantage of feature-level image fusion is its ability to deal with ignorance and missing information.

The MD classifier produced results very similar to that of the ML classifier and they have the least accurate of all according to experimental result of this study. Because they use a parametric rule that requires data normal distribution and well defined covariance's for each band in image data and each training class. Out of all four supervised classifiers ED Classifier generate more accurate classification results than other classifiers do, despite the popular belief that the ML classifier is the most accurate classifier.

**Acknowledgment**
The authors would like to thank DigitalGlobe for providing the data sets used in this paper.

## AUTHORS

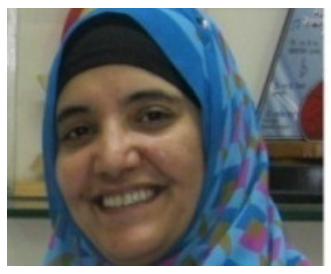

Firouz Abdullah Al-Wassai received her B.Sc. degree in Physics in 1993from University of Sana'a, Yemen, Sana'a and M.Sc.degree in Physics in 2003from Bagdad University, Iraq. Currently she is Research student PhD in the department of computer science (S.R.T.M.U), Nanded, India. She has published papers in twelve International Journals and conference.

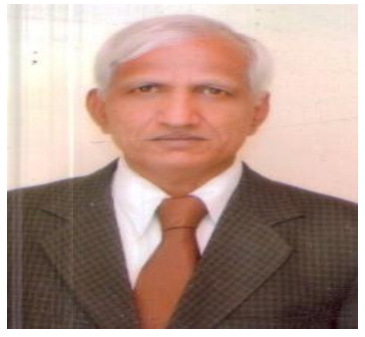

Dr. N.V. Kalyankar, Principal,Yeshwant Mahvidyalaya, Nanded(India) completed M.Sc.(Physics) from Dr. B.A.M.U, Aurangabad. In 1980 he joined as a leturer in department of physics at Yeshwant Mahavidyalaya, Nanded. In 1984 he completed his DHE. He completed his Ph.D. from Dr.B.A.M.U. Aurangabad in 1995. From 2003 he is working as a Principal to till date in Yeshwant Mahavidyalaya, Nanded. He is also research guide for Physics and Computer Science in S.R.T.M.U, Nanded. 03 research students are successfully awarded Ph.D in Computer Science under his guidance. 12 research students are successfully awarded M.Phil in Computer Science under his guidance He is also worked on various boides in S.R.T.M.U, Nanded. He is also worked on various bodies is S.R.T.M.U, Nanded. He also published 30 research papers in various international/national journals. He is peer team member of NAAC (National Assessment and Accreditation Council, India ). He published a book entilteld "DBMS concepts and programming in Foxpro". He also get various educational wards in which "Best Principal" award from S.R.T.M.U, Nanded in 2009 and "Best Teacher" award from Govt. of Maharashtra, India in 2010. He is life member of Indian "Fellowship of Linnean Society of London(F.L.S.)" on 11 National Congress, Kolkata (India). He is also honored with November 2009.